\documentclass[a4paper]{cas-sc}

\usepackage{hyperref}
\usepackage{subfigure}

\usepackage[numbers]{natbib}
\usepackage{soul}
\usepackage{color}
\usepackage{float}
\usepackage{lineno}

\usepackage{textcomp}

\def\tsc#1{\csdef{#1}{\textsc{\lowercase{#1}}\xspace}}
\tsc{WGM}
\tsc{QE}

\begin{document}
\let\WriteBookmarks\relax
\def\floatpagepagefraction{1}
\def\textpagefraction{.001}

\shorttitle{}
\title [mode = title]{Hearing Your Blood Sugar: Non-Invasive Glucose Measurement Through Simple Vocal Signals, Transforming any Speech into a Sensor with Machine Learning}

\author[1]{Nihat Ahmadli}[orcid=0009-0005-2008-4533]

\affiliation[1]{organization={Electronics and Communication Engineering Department, Istanbul Technical University},
            addressline={Maslak}, 
            city={Istanbul}, 
            country={Turkey}}
            
\author[1]{Mehmet Ali Sarsil}[orcid=0009-0000-0991-0578]

\author[1]{Onur Ergen}[orcid=0000-0001-7226-4898]
\ead{oergen@itu.edu.tr}
\cormark[1]
\cortext[1]{Corresponding author. Department of Electronics and Communication Engineering, Istanbul Technical University Faculty of Electrical and Electronics, Sariyer, Istanbul, 34469, Turkey}


\begin{abstract}
Effective diabetes management relies heavily on the continuous monitoring of blood glucose levels, traditionally achieved through invasive and uncomfortable methods. While various non-invasive techniques have been explored—such as optical, microwave, and electrochemical approaches—none have effectively supplanted these invasive technologies due to issues related to complexity, accuracy, and cost. In this study, we present a transformative and straightforward method that utilizes voice analysis to predict blood glucose levels. Our research investigates the relationship between fluctuations in blood glucose and vocal characteristics, highlighting the influence of blood vessel dynamics during voice production. By applying advanced machine learning algorithms, we analyzed vocal signal variations and established a significant correlation with blood glucose levels. We developed a predictive model using artificial intelligence, based on voice recordings and corresponding glucose measurements from participants, utilizing logistic regression and Ridge regularization. Our findings indicate that voice analysis may serve as a viable non-invasive alternative for glucose monitoring. This innovative approach not only has the potential to streamline and reduce the costs associated with diabetes management but also aims to enhance the quality of life for individuals living with diabetes by providing a painless and user-friendly method for monitoring blood sugar levels.
\end{abstract}




\maketitle
\section{Introduction}\label{intro}
Diabetes mellitus (DM) is characterized as a metabolic disorder, marked by elevated blood glucose levels surpassing 230 mg/dL, a condition referred to as hyperglycemia, or a decrease below 65 mg/dL, identified as hypoglycemia \cite{review1}. Individuals afflicted with diabetes experience an impairment in either the production or effective utilization of the hormone insulin. Insulin, a pivotal regulator of glucose, engages with insulin receptors, facilitating the absorption of glucose by cells to serve as an energy source \cite{insulin}. Prolonged diabetes is associated with chronic complications, including but not limited to heart disease, kidney disease, stroke, vision impairment, and damage to the nervous system \cite{chronic}.

The worldwide prevalence of diabetes is substantial and steadily increasing. Projections by the World Health Organization (WHO) suggest that the number of individuals aged 18 to 99 with diabetes will reach 693 million by 2045, a notable surge from the reported 451 million in 2017 \cite{review2}. Within the United States, there is a projected 54\% escalation in the diabetic population, with estimates indicating a rise from 35.6 million individuals in 2015 to surpass 54.9 million by the year 2030 \cite{rev3}. Furthermore, diabetes places a substantial economic burden on individuals inflicted with the disease, as well as on healthcare providers and employers. On a global scale, diabetes places a considerable economic burden on patients, their families, healthcare systems, and national economies. This stems from direct treatment expenses and indirect costs related to productivity loss and decreased wages \cite{rev4}. 
For better diabetes control and treatment, frequent monitoring, convenience of blood glucose measurement, precision, and real-time measurement are essential. Today, the most accurate technique for measuring blood glucose is self-monitoring blood glucose using the traditional finger prick method \cite{rev5}. However, individuals who must monitor their blood glucose levels multiple times a day may find this procedure to be uncomfortable, inconvenient, and prone to infection, which renders the method invasive \cite{rev5, rev6}.

Thus, several continuous and non-invasive ways of measuring glucose levels have been proposed over the last decade \cite{rev7, cont2, breath, ecg2, noninv}. To illustrate, continuous glucose monitoring (CGM) technology has become a cutting-edge technique to improve diabetes care within the last ten years \cite{cgm_2}. Unlike conventional glucose meters, which provide a single blood glucose reading at the time of testing, continuous glucose monitoring (CGM) provides semi-continuous glucose level information. This is accomplished indirectly by using a specialized algorithm to extrapolate blood glucose levels from interstitial fluid glucose \cite{rev7, cgm_2}. However, CGMs have limitations associated with cost, accuracy, semi-invasiveness, and the need for calibration after certain periods \cite{rev8}.

In the literature, voice has been extensively employed as a diagnostic biomarker, as the voice can be affected by various diseases and pathological circumstances because of transient or static alterations in the speech organs of the speaker or in the brain mechanisms that control speech \cite{alzheim, rev10, heart, rev9, depression, voice_3, ahmadlivoice}. In \cite{alzheim}, authors used voice data to distinguish between normal cognition and cognitive impairment (CI) or Alzheimer’s disease dementia (ADD) by extracting voice features such as Mel-frequency cepstral coefficients and Chroma, employing a deep neural network (DNN) model that achieved an accuracy of approximately 82\% in predicting ADD. In \cite{rev10}, researchers investigated the use of voice source information to detect Parkinson's disease (PD) by extracting glottal features through iterative adaptive inverse filtering and quasi-closed phase glottal inverse filtering methods. They compared traditional pipeline approaches and end-to-end deep learning models, finding that both methods provided modest classification accuracies, with the highest accuracy of 68.56\% achieved using a deep learning model trained on QCP-based glottal flow signals. Additionally, researchers in \cite{heart} developed artificial neural networks (ANNs) to recognize vocal distortions caused by heart failure (HF). Voice features were extracted using techniques such as discrete wavelet transform, fast Fourier Transform, and Mel-cepstral analysis. The ANNs achieved an efficiency of 96.7\%, demonstrating the potential of voice analysis in HF recognition.

\begin{figure*}[t]
	\centering
		\includegraphics[width=0.8\linewidth, height=0.50\linewidth]{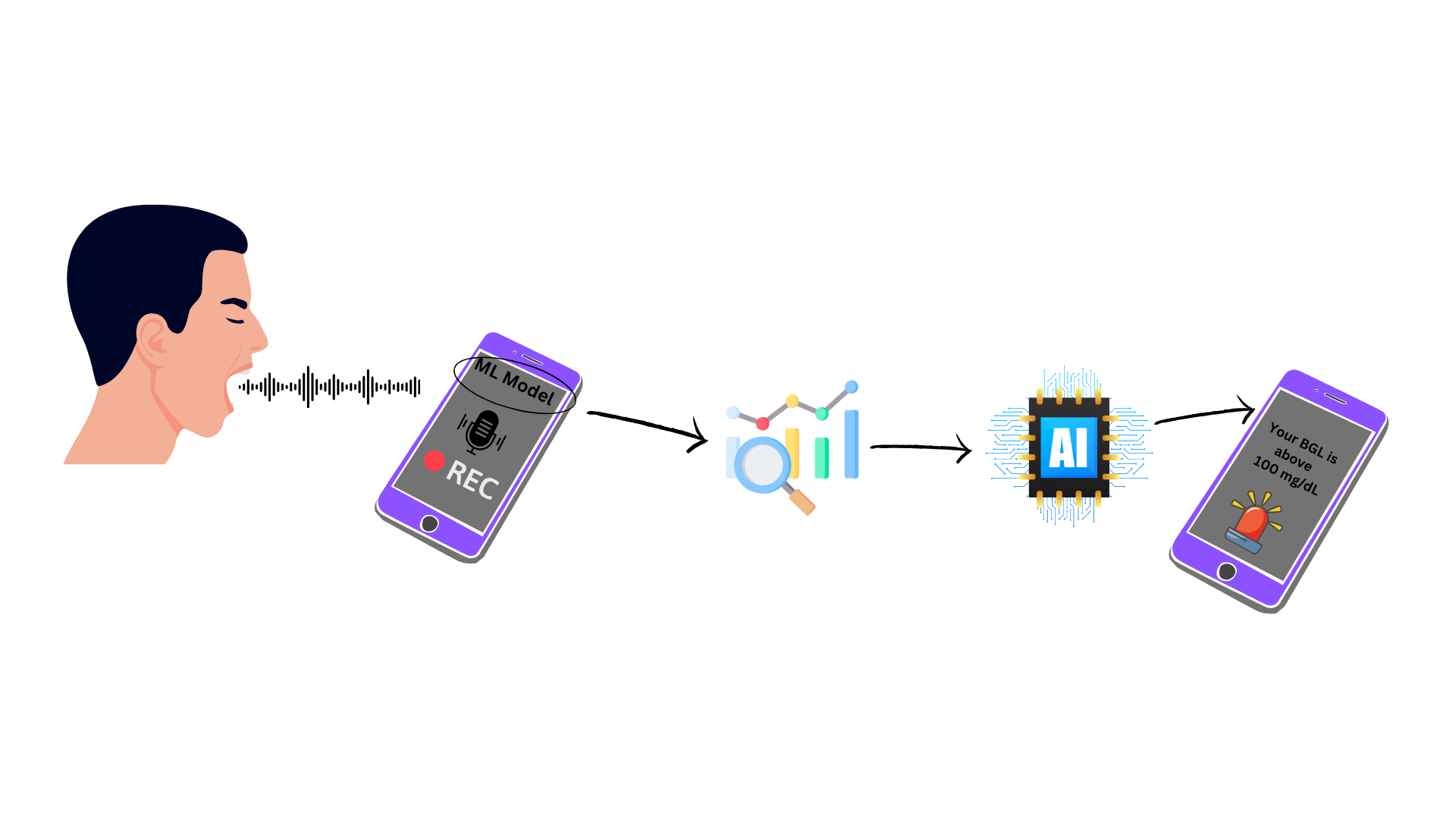}
	  \caption{The simplified depiction of our approach}\label{fig: summary}
\end{figure*}

Voice synthesis is a complex process that involves the respiratory system, nervous system, and larynx. Any disruptions in these systems can alter the voice, either in ways that are audibly noticeable or detectable through computer analysis \cite{mechanics}. Fluctuations in blood glucose levels can alter the elasticity of the vocal cords and the laryngeal soft tissue, affecting tissue compliance based on Hooke’s law of physics \cite{glucovoicerev}. Furthermore, chronic high or low glucose can lead to complications like peripheral neuropathy \cite{neuropathy} and myopathy \cite{myopathy}, which involve damage to nerves and muscle fibers in virtually every part of the body, including those associated with vocal fold. Myopathy has been linked to a higher incidence of voice disorders and difficulty swallowing, possibly due to muscle weakness in the larynx \cite{dys}, while conditions such as hoarseness, vocal strain, and loss of voice are often seen in those with diabetic neuropathy\cite{pho}. Additionally, research in voice acoustic analysis has revealed notable differences in fundamental frequency, jitter, shimmer, and noise-to-harmonic ratio between healthy individuals and those with diabetes \cite{hamdan2012vocal, instrumental}.

Following this, recent studies have highlighted the potential of using vocal biomarkers for non-invasive glucose monitoring, showing the correlation between these biomarkers and fluctuating glucose levels \cite{cfrd, acousticgluco, voicediabete, patent}. For instance, research in \cite{cfrd} has demonstrated that the noise-to-harmonic ratio (NHR) and fundamental frequency variation (vF0) can significantly indicate glycemic control and high blood glucose levels in patients with cystic fibrosis-related diabetes (CFRD), suggesting the feasibility of voice analysis as a diagnostic tool. Furthermore, another study in \cite{acousticgluco} utilized smartphone applications to record and analyze voice samples from participants, extracting features such as pitch, jitter, and shimmer to accurately predict Type 2 Diabetes Mellitus (T2DM). Findings in \cite{voicediabete} suggest that both hypoglycemia and hyperglycemia significantly modulate human voice characteristics, with distinct patterns observed between genders. This suggests the potential for using voice analysis as a non-invasive method for early detection and prevention of hypoglycemia in people with diabetes. There is even a patent that detects hypoglycemia and hyperglycemia through acoustic-phonetic analysis of speech during phone calls, comparing the current speech to baseline recordings of the speaker's normal state \cite{patent}.

In this study, we developed a machine learning-based blood glucose estimation tool designed to classify blood glucose levels in individuals as high and low. We conducted data collection among volunteer individuals, who participated in the speech protocol by providing their audio records following a transcript and measuring their blood glucose levels by a glucometer. Specifically, logistic regression with Ridge regularization was employed for training the machine learning model with a threshold of 100 mg/dL for classification into high (>100mg/dL) and low (<100mg/dL) blood glucose levels. The rationale for our choice is that a threshold of 100 mg/dL for classifying blood glucose levels as high or low is a well-established standard in diabetes management \cite{american20212, who}. Our model's accuracy and reliable performance, validated through cross-validation and rigorous statistical methods (p-value $<$ 0.01), positions it as further proof for the integration of voice analysis and AI in non-invasive blood glucose level measurement. To our best knowledge, this is the first study to utilize Machine Learning to assess changes in statistical vocal parameters associated with high and low blood glucose levels. The simplified depiction of our study is shown in Fig. \ref{fig: summary}

\section{Materials and methods}
\subsection{Data Collection}
The data was collected from 49 participants, including both their vocal recordings and corresponding blood glucose levels. 6 of these volunteers have type-1 diabetes. The data collection spanned over 4 weeks and comprised a cohort of 30 males and 19 females. Exclusion criteria were established to eliminate individuals with conditions affecting text readability, such as significant respiratory distress, as well as those with voice quality-impacting malignancies, fever, smoking habit, or recent vaccination (within the past week). Additionally, individuals unable to provide informed consent and those with articulation disorders were excluded from the study. Some baseline characteristics of the study cohort are provided in Table \ref{Table: Grid}.

In individuals with normal physiological metabolism, plasma glucose concentrations are carefully regulated within a narrow range throughout the day. Typically, these concentrations average between 70 and 100 mg/dl after an overnight fast and before meals. Postprandially, or after meals, the levels generally do not exceed 160 mg/dl \cite{glucose}.

Each participant was requested to contribute data at least once and at most twice to the dataset, one with a high glucose level and one with a low glucose level, resulting in a total of 70 voice samples. If a participant provided a data sample during a period of lower blood glucose levels, a glucose challenge test was administered to elevate the blood glucose level to the upper extremity. This test, resembling a condensed version of the glucose tolerance test, could be conducted at any time of the day. It involved the consumption of a glass of concentrated glucose solution (50 g of glucose dissolved in 250 to 300 ml of water). After one hour, a blood sample was drawn to determine the blood sugar level \cite{oralglucose}.

\begin{table}[H]
\centering
    \caption{\label{Table: Grid} Basic Characteristics of Participants}
    \small\addtolength{\tabcolsep}{2.5pt}
    \begin{tabular}{| l | l | l | l |}
    \hline
    & Blood Glucose Level $<$ 100 mg/dl & Blood Glucose Level $>$ 100 mg/dl \\
      \hline
     Age  & $22.83 \pm 3.04$ &  $22.95 \pm 3.85$ \\
     \hline
     Sex(male) & 13(57.1) & 15(58.3) \\ 
     \hline
     Measured BG (mg/dL)  & $81.3 \pm 15.28$ & $169 \pm 51.24$ \\
     \hline
     Total participants & 23 & 26 \\
     \hline
    \end{tabular}
\end{table}

\subsection{CAPE-V Protocol and Voice Recordings} \label{sub: capev}
The speech protocol employed in this study follows the CAPE-V Protocol, which stands for Consensus Auditory-Perceptual Evaluation of Voice. The CAPE-V  was developed as a clinical tool for auditory perceptual voice assessment \cite{capev}. Its primary aim is to characterize the auditory-perceptual features associated with a voice issue, facilitating effective communication among healthcare professionals. The protocol was formulated by the American Speech-Language-Hearing Association’s (ASHA) Special Interest Division 3, Voice and Voice Disorders, and achieved consensus adoption at a conference held in 2002 at the University of Pittsburgh \cite{capev}.

The assessment involves gathering voice samples from patients across four sections:

\begin{itemize}
\item Six sentences specifically crafted based on different phonetic contexts.
\item Sustained vowel "a" for 5 seconds.
\item Sustained vowel "i" for 5 seconds.
\item Conversational speech, including responses to at least two of three daily questions.
\end{itemize}

Patients are instructed to record their voices, following the Turkish version of the CAPE-V protocol \cite{capev_turk}, which includes the four sequential segments mentioned above. The average duration of each segment collected is 31.3, 4.4, 5.2, and 37.0 seconds, respectively. To ensure standardized and clear recordings, the cell phone was positioned 20-30 cm from the volunteer's mouth, and recordings were made in a quiet environment to minimize background interference.

To construct a supervised machine learning model and necessitate labels, we designated the target variable in our classification problem. For the purpose of binary classification, a predefined threshold of 100 mg/dL for blood glucose levels was set. Individuals with blood glucose levels exceeding this threshold were labeled as 1, indicating high blood glucose levels, while those below it were labeled as 0, indicating normal-to-low blood glucose levels.

\begin{table*}[h] 
\caption{\label{Table: disvoice_param} Vocal parameters and their descriptions}
    \centering
    \begin{tabular}{| p{1.5cm} | p{3.5cm} | p{1.2cm} | p{3.2cm} | p{1.8cm} | p{3.5cm} |}
        \hline
        \multicolumn{2}{|c|}{Glottal Features}
        & \multicolumn{2}{c|}{Phonation Features}
        & \multicolumn{2}{c|}{Prosody Features} \\
        \hline
        GCI & Time variability between glottal closure instances
        & F'0 & The rate of pitch change in a voice signal
        & Tilt of linear estimation of F0 for each voiced segment
        & The rate of change or slope in the pitch (fundamental frequency) over each voiced segment \\
         \hline
        Average of OQ & Mean opening quotient for sequential glottal cycles, indicating the ratio of the opening phase duration to the total cycle duration
        & F''0 & The rate of change in the rate of pitch change in a voice signal & Energy on the first segment & The initial intensity or loudness of the voiced portion of speech \\
        [15.5ex] \hline
        Variability of OQ & Variation in the opening quotient for sequential glottal cycles, describing the consistency in the opening phase duration relative to the total cycle duration & Jitter & Irregularity in the timing between consecutive pitch cycles in a voice & Energy on last segment & The loudness level at the end of a voiced speech segment \\
        [16.5ex] \hline
        Average of NAQ	& Mean Normalized Amplitude Quotient for sequential glottal cycles, showing the ratio of the amplitude quotient to the total cycle duration & Shimmer & Variation in the amplitude of successive pitch cycles in a voice signal & Voiced rate & The frequency of voiced segments per second \\
        [15.5ex] \hline
    \end{tabular}
\end{table*}

\subsection{Data Preprocessing}
\subsubsection{Feature Extraction}
In our data processing pipeline, the 'Disvoice' library in the Python programming language played a crucial role in extracting relevant vocal parameters from participant voices. This framework computes glottal, phonation, and prosody features from raw speech files \cite{disvoice1, disvoice2, disvoice3, disvoice4}. For glottal features, Disvoice focuses on evaluating nine descriptors, including the "Average Harmonic Richness Factor (HRF)." This descriptor calculates the average ratio between the sum of the amplitudes of harmonics and the amplitude of the fundamental frequency, providing insights into the richness of harmonic content in the voice \cite{disvoice1}. The phonation features computed by the model reveal abnormal and impaired patterns in vocal fold vibration, assessed in terms of stability measures. Disvoice predominantly concentrates on evaluating seven descriptors, encompassing parameters such as Jitter and Shimmer. Jitter quantifies variations in the fundamental period of the voice signal, while Shimmer assesses variations in voice signal amplitude \cite{disvoice3}. Prosody features encapsulate aspects of speech conveying information about timing, pitch variations, and volume levels in natural speech. Disvoice emphasizes the computation of 103 descriptors, including the  F0-contour parameter capturing pitch fluctuations over time, providing insights into the rhythmic, intonational, and emotional aspects of speech \cite{disvoice4}. Vocal features are computed on frames obtained by dividing the audio signals. Additionally, calculations of statistical parameters—average, standard deviation, maximum, minimum,  skewness, and kurtosis—over the vocal features across the frames were included in the extracted features. Table \ref{Table: disvoice_param} provides examples of vocal parameters derivable from the three feature sets. In our feature extraction process, glottal features were exclusively derived from sustained vowels due to the distinctive characteristics of sustained vowels. Sustained vowels provide a controlled and stable environment,  facilitating a focused study of glottal behavior \cite{disvoice4}. Phonation and prosody features,  on the other hand, were extracted from all segmented voice sections mentioned in  Section \ref{sub: capev} Prosody features were extracted from conversational speech segments, as conversational speech might include spontaneous pauses related to the thought process. Extracting these features from conversational speech could yield data regarding the mental influence of blood glucose levels on an individual.

\subsection{Feature Engineering}
\begin{figure*}[t]
	\centering
		\includegraphics[width=1.0\linewidth, height=0.45\linewidth]{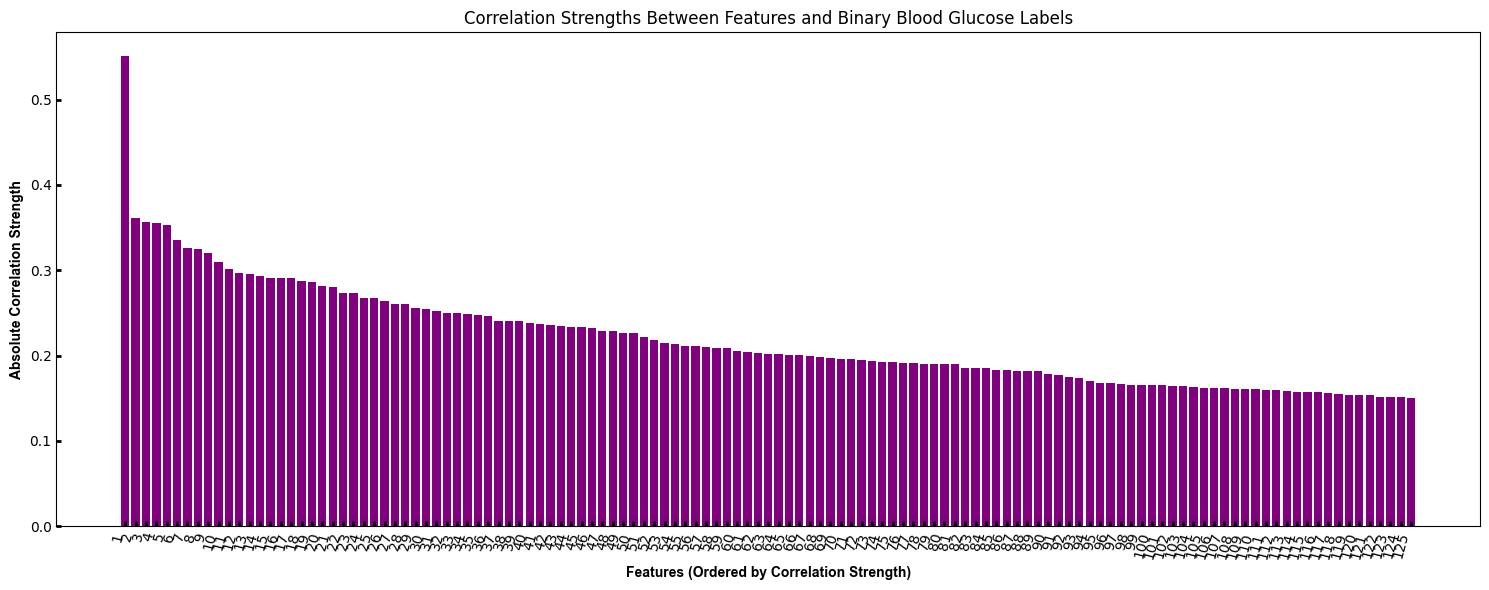}
	  \caption{Correlation of remaining features with target variable}\label{fig: correlation}
\end{figure*}

A total of 596 features were extracted from each participant across all speech segments using the Disvoice library. Subsequently, these features underwent a reduction process, leading to the selection of 124 features. This reduction was achieved by applying a correlation threshold of 0.15 (equivalent to 15\%) with the target variable.  Specifically, features with a correlation below 15\% with the target variable were excluded, while the remaining 124 features were retained. The chosen correlation threshold aimed to prioritize features with a stronger relationship to the target variable,  focusing on the most informative and predictive features while eliminating those with weaker associations. The correlation among the final set of 124 features is visually depicted in Figure \ref{fig: correlation}. 

\begin{figure}[h]
	\centering
		\includegraphics[width=0.8\linewidth, height=0.50\linewidth]{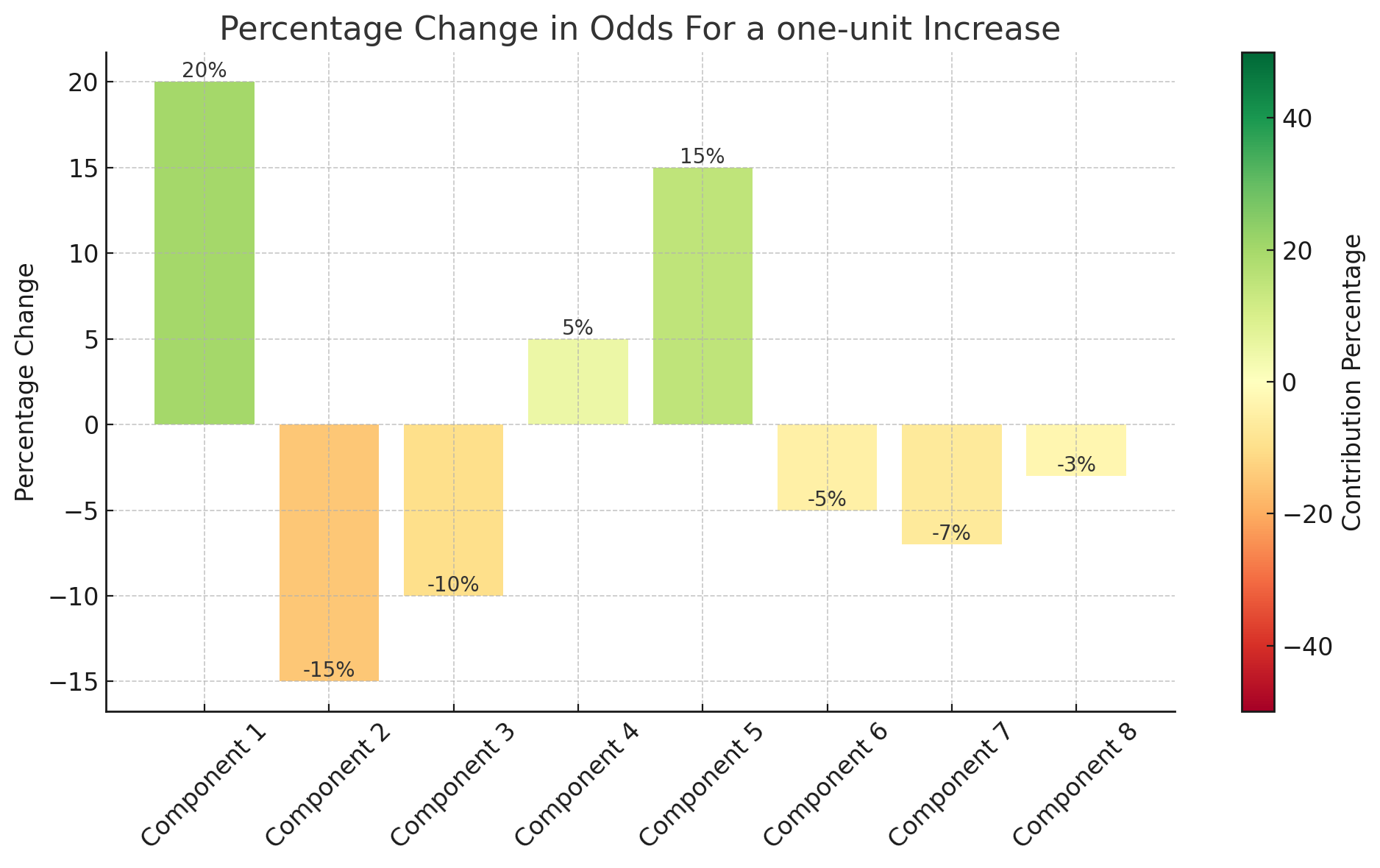}
	  \caption{Contribution Percentage of each Principal Component}\label{fig: contribution}
\end{figure}

To address challenges with the feature set, several pre-processing steps were implemented to enhance the effectiveness of the machine learning model. Firstly,  given the high dimensionality of the dataset with 124 features, susceptibility to the  "curse of dimensionality" was acknowledged \cite{curse}. This phenomenon could make the machine learning model prone to overfitting, increase computational intensity,  and necessitate larger datasets for effective generalization. Secondly, the presence of collinearity among most of the 124 features posed a risk of model instability,  hindered interpretability, and increased the likelihood of overfitting, collectively impacting the accuracy and reliability of predictions. To mitigate these issues,  Principal Component Analysis (PCA), a feature dimensionality reduction technique, was applied. PCA is an unsupervised learning method that utilizes patterns within high-dimensional data to simplify data complexity while retaining essential information \cite{pca1}. PCA performs a linear mapping of the data to a lower-dimensional space containing linearly uncorrelated features, maximizing the variance of the data in the low-dimensional representation while preserving maximum information. In our model, PCA reduced the initial dataset of 124  features by identifying and retaining 8 linear combinations of these features,  known as principal components. These principal components capture the most significant variance in the data and, importantly, are linearly uncorrelated with each other \cite{pca1}. Examining the loading scores of features that contributed to the resulting components provides insight into the constitution of these components.  The 8 principal components, serving as inputs to the devised algorithm, along with their contribution percentage to the correct prediction of the model, are illustrated in Fig. \ref{fig: contribution}. These preprocessing steps facilitated the creation of a more interpretable and robust model, allowing for meaningful insights and conclusions.

\section{Model Development and Training} \label{sec: model}
Given the binary classification nature inherent in our research task, we have chosen  Logistic Regression (LR) as the most suitable modeling approach \cite{logistic}. In the field of machine learning and statistics, LR serves as a predictive analytical method grounded in the fundamental concepts of probability. Its application is particularly pertinent in decision-making scenarios characterized by two distinct choices, such as the diagnosis of medical conditions or the prediction of a patient's risk. LR  operates by analyzing datasets, identifying discernible patterns, and ultimately making determinations between two possible outcomes. For example, it can ascertain whether a patient is afflicted with a specific ailment (affirmative or negative). In the context of our specific inquiry, LR is employed to predict whether a participant's blood glucose level surpasses a predefined threshold. 

Within the framework of Logistic Regression (LR), our initial step involves generating a linear combination utilizing the input features X, comprising 8  principal components in our specific case. This linear combination incorporates a set of coefficients ($\theta$) denoting the weights of each feature plus an intercept ($\theta_0$):

\begin{equation} \label{eq: z}
    \centering
    z=\ \theta_0\ +\ \theta_1.x_1\ +\ \theta_2.x_2\ +\ \ \ldots.\ {\ +\ \ \theta}_n.x_n 
\end{equation}

\begin{table*}[t]
\centering
    \caption{\label{Table: cv} The optimal hyperparameters chosen by GridSearchCV}
    \small\addtolength{\tabcolsep}{2.5pt}
    \begin{tabular}{| p{3.5 cm} | p{3.5 cm} | p{8.5cm} |}
    \hline
     Hyperparameter & GridSearchCV output & Explanation \\
     \hline
     “C”	& 0.9 & This parameter represents the inverse of the regularization strength. A lower "C" value indicates stronger regularization, while a higher "C" value weakens the regularization effect. \\
     [6.0ex]\hline
     “penalty” & “L1” & This parameter works together with "C" to define the type of regularization applied. L2 regularization typically retains all features but reduces their coefficients toward zero. \\
     [9.0ex]\hline
     “solver” & “liblinear”	& This parameter selects the optimization algorithm used to train the logistic regression model. "liblinear" is an efficient solver for binary classification, especially with small datasets, and it works well with both L1 and L2 regularization. \\ 
     [11.5ex]\hline
    \end{tabular}
\end{table*}

where $z$ is the linear combination, $\theta_0$ is the intercept (also known as the bias term), and $\theta_i$ are the weights (coefficients) for the $i^th$ feature. This linear combination, $z$, represents a measure of evidence for the positive class (class 1).

Once fitted to a dataset, LR predicts the probability of a positive class $p(y_i=1 | X_i)$ with the help of a logistic function (also called sigmoid function) \cite{sigmoid}:

\begin{equation} \label{eq: sigmoid}
\ p=\ \frac{1}{1+\ e^{-(\theta_0\ +\ \theta_1.x_1\ +\ \theta_2.x_2\ +\ \ \ldots.\ {\ +\ \ \theta}_n.x_n)}}\ 
\end{equation}

where e is the base of the natural logarithm and p is the probability of belonging to class 1 (glucose level > 100 mg/dL).

Subsequently, to make a classification decision, a threshold (often set at 0.5) is applied to the predicted probability $p$. If $p$ is greater than or equal to this threshold, the data point is classified as belonging to class 1; otherwise, it is classified as class 0.

To effectively train the model, optimal values for the coefficients $\theta$ need to be determined. This is typically achieved through a process known as Maximum Likelihood Estimation \cite{mle}. The objective is to identify the values of $\theta$ that maximize the likelihood of the observed data given the model. Afterward, a cost or loss function \cite{loss} is employed to quantify the error between the predicted probabilities and the true class labels in the training data:

\begin{equation}
\mathcal{L}(y_i, p_i) = \frac{1}{m}\sum_{i=0}^{m} [y_i \cdot \log(p_i)
+ (1-y_i) \cdot \log(1-p_i)] 
\label{eq:loss}
\end{equation}

where $m$ is the number of training data samples, $y_i$, and $p_i$ are the actual class label (0 or 1) and predicted probability of class 1 for the $i^th$ sample, respectively.

The optimization algorithm then seeks to minimize this loss function by updating the coefficients $\theta$ in each iteration until the loss converges to a minimum value.

The primary distinction between Logistic Regression (LR) in machine learning and its counterpart in statistics lies in the incorporation of regularization. Regularization serves the purpose of preventing overfitting by introducing a penalty term to the model's cost function \cite{regularization}, which assesses the model's fit to the data. Binary class logistic regression with the addition of a regularization term aims to minimize the cost function in Eq. \ref{eq:loss}. Numerous regularization techniques with the penalization arguments, such as $l1$ and $l2$, exist. The penalization argument can be regarded as a hyperparameter of the LR model, a parameter that is used to configure an ML model \cite{hyperparameter}. 

\begin{figure*}[t] 
	\centering
 \begin{subfigure}{}
    \includegraphics[width=0.40\linewidth,height= 0.4\linewidth]{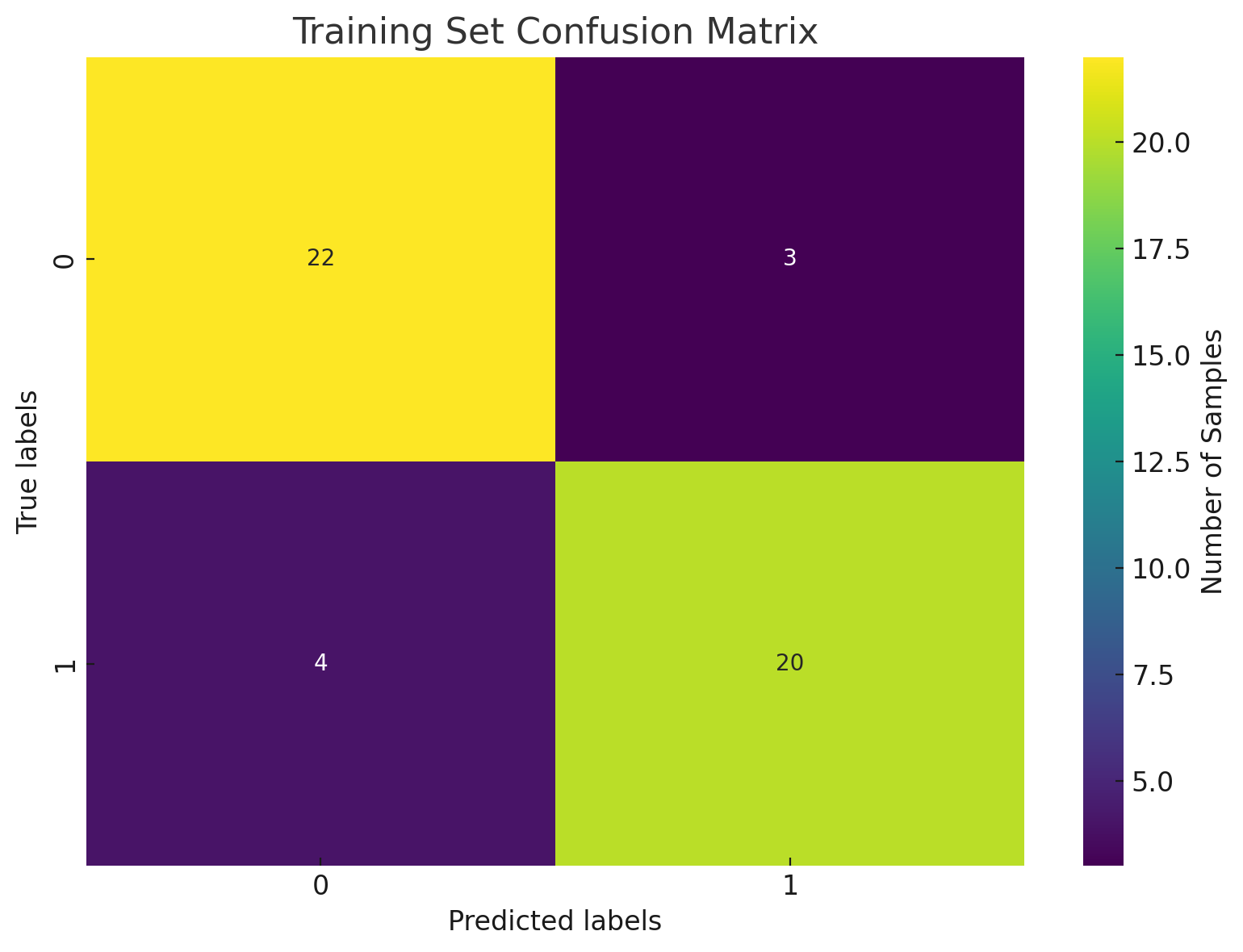}
    \end{subfigure}
    \begin{subfigure}{}
    \includegraphics[width=0.40\linewidth,height=0.40\linewidth]{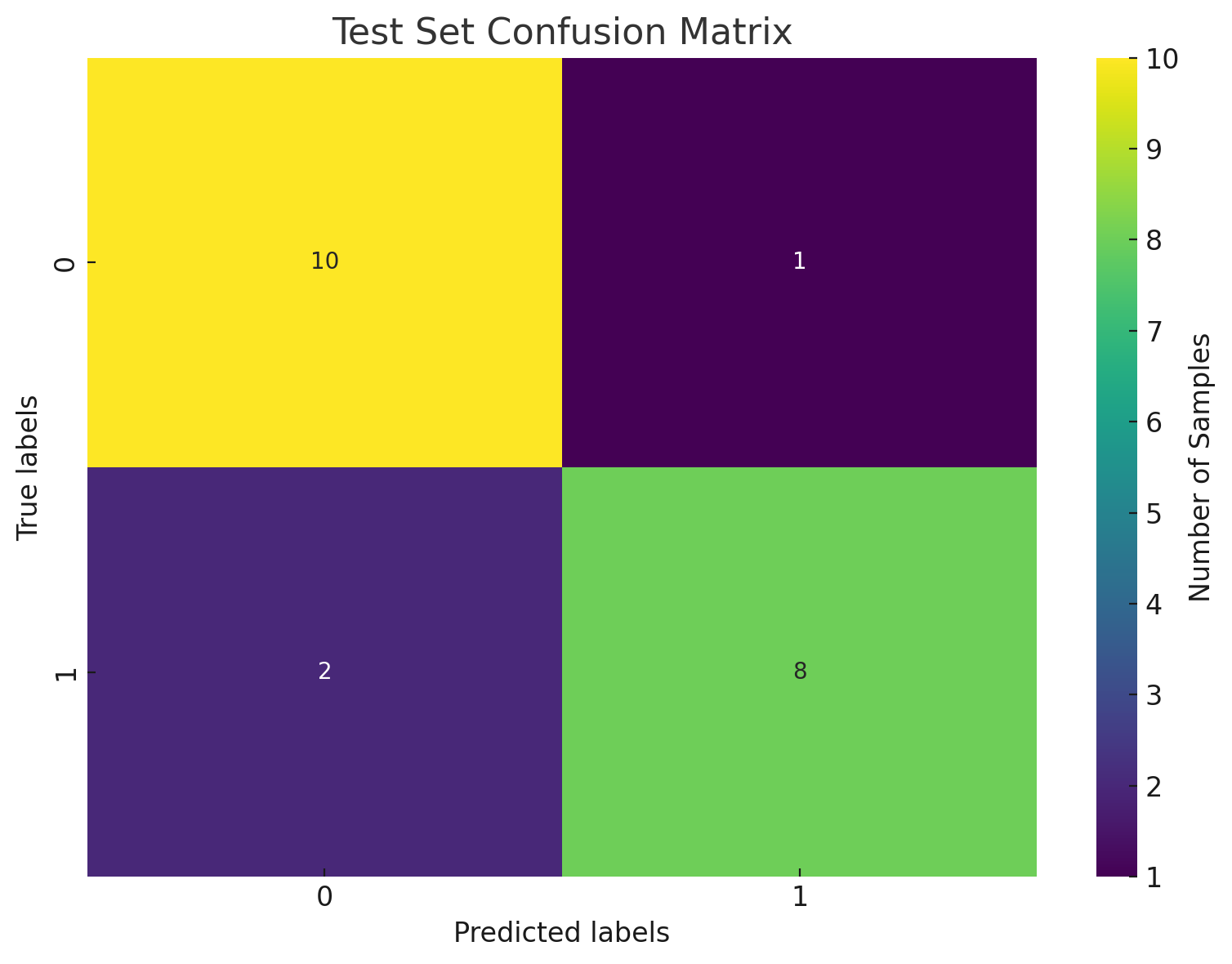}
    \end{subfigure}
	\caption{\label{fig: conf_mat1} Confusion matrix evaluated on the LR model applied on both splits}
\end{figure*}

To apply the model to our data, the dataset comprising 70 voice records was initially split into training and test sets, with a test ratio of 0.3 for model evaluation,  resulting in 49 training samples and 21 test samples. GridSearchCV \cite{gridsearch}, a hyperparameter tuning technique, was then utilized to systematically search and select optimal hyperparameters associated with the LR model. This involved an exhaustive search over a predefined hyperparameter grid, considering various combinations of hyperparameters, such as regularization strength and penalty type.  The selection criteria were based on performance metrics like accuracy, precision,  and recall, aiming to identify hyperparameters that yield the best classification results. After hyperparameter tuning, the logistic regression model,  equipped with its best-selected hyperparameters, was trained using the 49 training samples and evaluated with the 21 test samples. The optimal hyperparameters for  LR, as determined through GridSearchCV, along with their corresponding meanings, are presented in Table \ref{Table: cv}. 

\section{Results}

We refer to the variable 'z' in Eq.\ref{eq: z} as a unitless measure called the 'acoustic predictor.' This predictor is derived from a combination of eight non-collinear principal components, as detailed in Section \ref{sec: model}. We found a strong statistical significance for this predictor in distinguishing between high and low blood glucose levels (p < 0.001).

In this section, the outcomes of the predictive model designed for evaluating blood glucose levels in participants are presented. Firstly, an overview of the performance attained by the machine learning model trained on the dataset is presented.  Subsequently, a leave-one-out cross-validation (LOOCV) technique is employed as a validation approach to evaluate the model's capacity to generalize to unseen data. 

As previously indicated, the dataset was partitioned into training (N = 49 samples)  and test (N = 21 samples) sets, with a test ratio of 0.3. The emphasis was on pivotal evaluation metrics, encompassing accuracy, precision, recall, and F1-score. The algorithm exhibited performance metrics for both the training and test sets, as elaborated in Table \ref{Table: results_train}. The confusion matrix assessed on the implemented logistic regression algorithm is depicted in Figure \ref{fig: conf_mat1}.

\begin{table}[h]
\centering
    \caption{\label{Table: results_train} Results of LR on Train-Test Split}
    \begin{tabular}{| p{ 1.8 cm} | p{1.1 cm} | p{1.1cm} | p{1.0cm}| p{1.2cm}|}
    \hline
    	& Accuracy & Precision	& Recall & F1-Score \\
     \hline
Training Set & 87.1\% & 0.86 & 0.95 & 0.90 \\
\hline
Test Set & 85.7\% & 0.71 & 1 & 0.83 \\
\hline
    \end{tabular}
\end{table}

\begin{figure}[h]
	\centering
		\includegraphics[width=0.5\linewidth, height=0.45\linewidth]{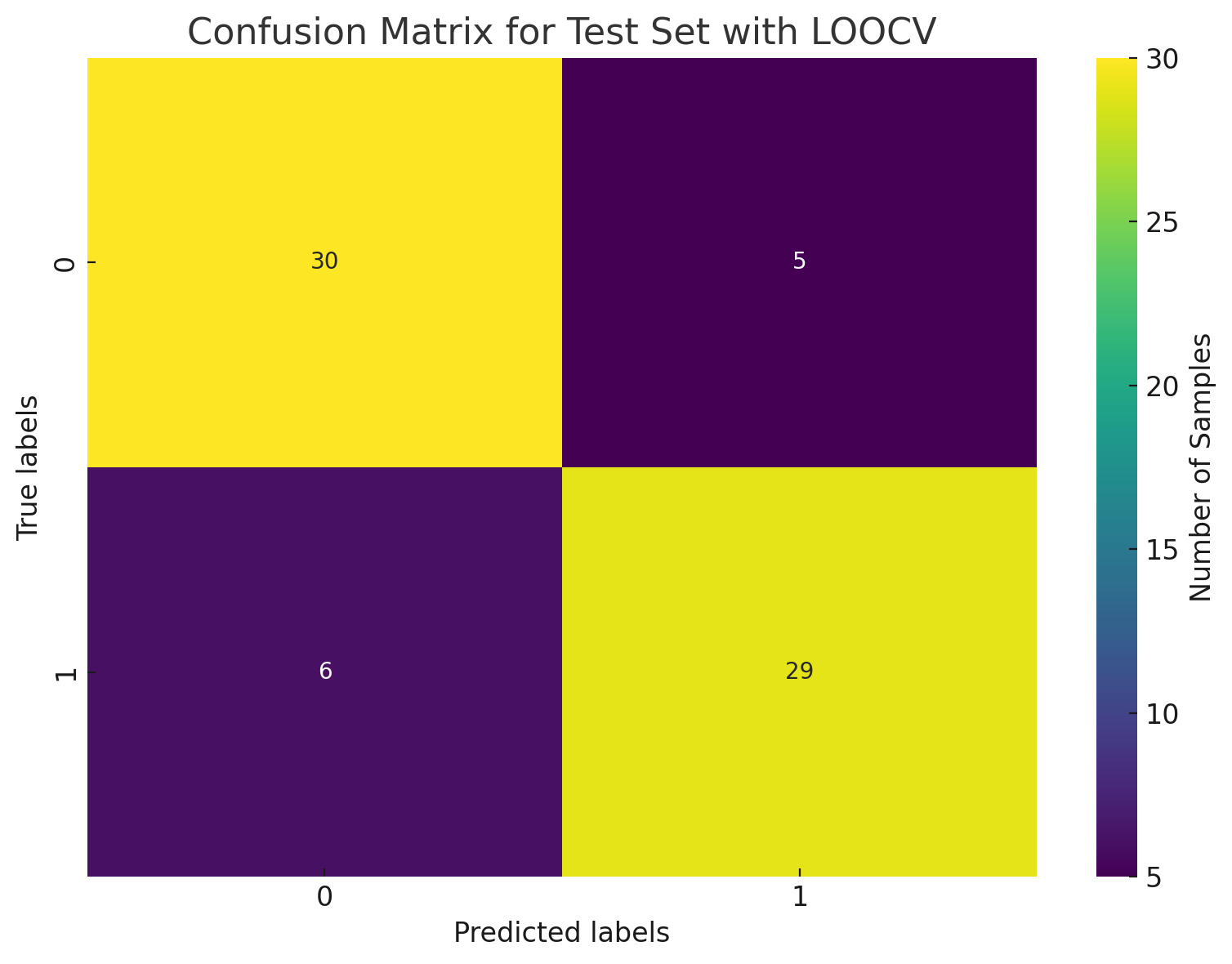}
	  \caption{Confusion matrix evaluated on LR with Leave-One-Out Cross-Validation}\label{fig: conf_mat2}
\end{figure}

To gauge the generalization capabilities of the developed model, a leave-one-out cross-validation (LOOCV) approach was employed on the entire dataset. This method is commonly adopted in studies with limited instances or class values,  ensuring a reliable accuracy estimate for a classification algorithm without the introduction of randomness in the instance division for training and testing \cite{loocv}.  Throughout this process, all 70 samples were iteratively cycled, with one sample excluded each time from the training set while the model was trained on the remaining data. This created a split ratio of 69 for training and 1 for testing,  resulting in a total of 70 unique combinations of training and test sets. Following the computation of individual evaluation metrics for the model on these 70  different training and test sets, overall metric values were derived by averaging the evaluation metrics for each combination over training and test sets separately. The results underscore the model’s robustness and its ability to generalize effectively across various test scenarios, establishing it as a promising tool for the targeted application \cite{loocv}. The averaged accuracy, precision, recall, and F1-score metrics are detailed in Table \ref{Table: results_loocv}. Additionally, the corresponding confusion matrix resulting from the test set predictions is depicted in Figure \ref{fig: conf_mat2}. 

\begin{table}[h]
\centering
    \caption{\label{Table: results_loocv} Parametric results of LR with Leave-One-Out Cross-Validation}
    \begin{tabular}{| p{ 1.8 cm} | p{1.1 cm} | p{1.1cm} | p{1.0cm}| p{1.2cm}|}
    \hline
    	& Accuracy & Precision	& Recall & F1-Score \\
     \hline
Training Set & 86.5\% & 0.83 & 0.94 & 0.88 \\
\hline
Test Set & 84.4\% & 0.81 & 0.81 & 0.86 \\
\hline
    \end{tabular}
\end{table}

\section{Discussion}
This study explores the potential of using vocal biomarkers as a non-invasive predictive tool for monitoring blood glucose levels, utilizing artificial intelligence (AI) techniques. The findings highlight a significant correlation between changes in voice features and blood glucose levels, suggesting that vocal analysis can be an effective and convenient method for glucose monitoring. The core idea is that changes in BGL can affect the vocal cords and associated tissues, altering the voice's physical properties. The study utilized a logistic regression model with Ridge regularization to classify BGL into high and low categories, based on a threshold of 100 mg/dL. Specifically features related to jitter was found to correlate with BGL because fluctuations in glucose levels can alter the elasticity of the vocal cords, leading to irregular vibrations. High jitter values indicate greater variability, which might correspond to higher BGL. Jitter refers to the cycle-to-cycle variations in the fundamental frequency (pitch) of the voice. It is a measure of the stability of vocal fold vibrations. 

The simplicity and non-invasive nature of this method present a substantial advantage over traditional glucose monitoring techniques, such as finger-pricking and continuous glucose monitoring systems (CGMs), which can be invasive, costly, and inconvenient. A fine-tuned logistic regression model is deployed for the training and testing of collected samples. Subsequently, the reproducibility and reliability of the results are evaluated through a cross-validation approach to assess the model’s generalization ability to unseen data. The model exhibits consistent results in both cases, dropping nearly 1\% in accuracy in cross-validation. Additionally, the acoustic predictor generated by the devised machine learning model demonstrates a significant correlation with sample labels (p $<$ 0.01), emphasizing the statistical significance of the relationship between the predictor and blood glucose levels. An inherent advantage of our study over current non-invasive blood glucose monitoring methods lies in the simplicity of our approach, necessitating only the individual’s voice and several easily obtainable physiological and environmental parameters as input.

While our study presents promising results, it is crucial to consider potential confounding factors that could influence the accuracy and reliability of vocal biomarkers for glucose monitoring. Stress, vocal strain, and various health conditions, such as respiratory infections or neurological disorders, can alter vocal characteristics independently of blood glucose levels. For instance, stress can lead to changes in pitch and vocal tension, while vocal strain may affect jitter and shimmer measurements. To mitigate these effects in future research, we recommend implementing more rigorous participant screening processes to exclude individuals with acute health conditions affecting vocal quality. Furthermore, longitudinal studies could be conducted to observe how consistent vocal biomarkers are over time and under varying physiological states. In addition to these, the relatively small sample size and homogeneity of the participant group may restrict the generalizability of the findings. Future research should aim to expand the dataset, include a more diverse participant pool, and explore the impact of these confounding factors.

\section{Conclusion}
In conclusion, this study demonstrates the promising potential of vocal biomarkers and artificial intelligence for non-invasive glucose monitoring. The high accuracy of the logistic regression model in predicting blood glucose levels from vocal features indicates that voice analysis could be seamlessly integrated into routine glucose monitoring practices. This innovation stands to significantly enhance the quality of life for diabetic patients by providing a painless and accessible method for managing their condition.

Beyond this proof-of-concept study, future research should prioritize the validation of these findings in larger and more diverse populations. Additionally, efforts should be directed towards refining the model’s accuracy and exploring its real-world applications. By advancing these initiatives, voice-based glucose monitoring could evolve into a standard tool in diabetes care, providing an innovative solution that leverages the capabilities of AI and vocal analysis. The successful implementation of this technology has the potential to revolutionize diabetes management, making it easier, more comfortable, and more cost-effective for millions of individuals worldwide.

\printcredits



\bibliographystyle{elsarticle-num} 
\bibliography{main}

\begin{thebibliography}{10}
\expandafter\ifx\csname url\endcsname\relax
  \def\url#1{\texttt{#1}}\fi
\expandafter\ifx\csname urlprefix\endcsname\relax\def\urlprefix{URL }\fi
\expandafter\ifx\csname href\endcsname\relax
  \def\href#1#2{#2} \def\path#1{#1}\fi

\bibitem{review1}
M.~Shokrekhodaei, S.~Quinones, Review of non-invasive glucose sensing techniques: optical, electrical and breath acetone, Sensors 20~(5) (2020) 1251.

\bibitem{insulin}
E.~Vargas, P.~Nandhakumar, S.~Ding, T.~Saha, J.~Wang, Insulin detection in diabetes mellitus: challenges and new prospects, Nature Reviews Endocrinology 19~(8) (2023) 487--495.

\bibitem{chronic}
D.~Mauricio, N.~Alonso, M.~Gratac{\'{o}}s, Chronic diabetes complications: the need to move beyond classical concepts, Trends in Endocrinology \& Metabolism 31~(4) (2020) 287--295.

\bibitem{review2}
H.~Sun, P.~Saeedi, S.~Karuranga, M.~Pinkepank, K.~Ogurtsova, B.~B. Duncan, C.~Stein, A.~Basit, J.~C. Chan, J.~C. Mbanya, et~al., Idf diabetes atlas: Global, regional and country-level diabetes prevalence estimates for 2021 and projections for 2045, Diabetes research and clinical practice 183 (2022) 109119.

\bibitem{rev3}
J.~da~Rocha~Fernandes, K.~Ogurtsova, U.~Linnenkamp, L.~Guariguata, T.~Seuring, P.~Zhang, D.~Cavan, L.~E. Makaroff, Idf diabetes atlas estimates of 2014 global health expenditures on diabetes, Diabetes research and clinical practice 117 (2016) 48--54.

\bibitem{rev4}
L.~Shi, V.~Fonseca, B.~Childs, Economic burden of diabetes-related hypoglycemia on patients, payors, and employers, Journal of Diabetes and its Complications 35~(6) (2021) 107916.

\bibitem{rev5}
A.~S. Bolla, R.~Priefer, Blood glucose monitoring-an overview of current and future non-invasive devices, Diabetes \& Metabolic Syndrome: Clinical Research \& Reviews 14~(5) (2020) 739--751.

\bibitem{rev6}
S.~A. Siddiqui, Y.~Zhang, J.~Lloret, H.~Song, Z.~Obradovic, Pain-free blood glucose monitoring using wearable sensors: Recent advancements and future prospects, IEEE reviews in biomedical engineering 11 (2018) 21--35.

\bibitem{rev7}
A.~Vazeou, Continuous blood glucose monitoring in diabetes treatment, Diabetes research and clinical practice 93 (2011) S125--S130.

\bibitem{cont2}
S.~Seidu, S.~K. Kunutsor, R.~A. Ajjan, P.~Choudhary, Efficacy and safety of continuous glucose monitoring and intermittently scanned continuous glucose monitoring in patients with type 2 diabetes: A systematic review and meta-analysis of interventional evidence, Diabetes Care 47~(1) (2024) 169--179.

\bibitem{breath}
H.~Liu, W.~Liu, C.~Sun, W.~Huang, X.~Cui, A review of non-invasive blood glucose monitoring through breath acetone and body surface, Sensors and Actuators A: Physical (2024) 115500.

\bibitem{ecg2}
M.~Andellini, R.~Castaldo, O.~Cisuelo, M.~Franzese, M.~S. Haleem, M.~Ritrovato, L.~Pecchia, R.~Schiaffini, Are the variations in ecg morphology associated to different blood glucose levels? implications for non-invasive glucose monitoring for t1d paediatric patients, Diabetes Research and Clinical Practice 212 (2024) 111708.

\bibitem{noninv}
P.~Jain, A.~M. Joshi, S.~P. Mohanty, L.~R. Cenkeramaddi, Non-invasive glucose measurement technologies: Recent advancements and future challenges, IEEE Access (2024).

\bibitem{cgm_2}
D.~A. Mihai, D.~S. Stefan, D.~Stegaru, G.~E. Bernea, I.~A. Vacaroiu, T.~Papacocea, M.~O.~D. Lupușoru, A.~E. Nica, O.~Stiru, D.~Dragos, et~al., Continuous glucose monitoring devices: A brief presentation, Experimental and therapeutic medicine 23~(2) (2022) 1--6.

\bibitem{rev8}
R.~Lin, F.~Brown, S.~James, J.~Jones, E.~Ekinci, Continuous glucose monitoring: a review of the evidence in type 1 and 2 diabetes mellitus, Diabetic Medicine 38~(5) (2021) e14528.

\bibitem{alzheim}
C.-Y. Park, M.~Kim, Y.~Shim, N.~Ryoo, H.~Choi, H.~T. Jeong, G.~Yun, H.~Lee, H.~Kim, S.~Kim, et~al., Harnessing the power of voice: A deep neural network model for alzheimer’s disease detection, Dementia and Neurocognitive Disorders 23~(1) (2024) 1.

\bibitem{rev10}
N.~Narendra, B.~Schuller, P.~Alku, The detection of parkinson's disease from speech using voice source information, IEEE/ACM Transactions on Audio, Speech, and Language Processing 29 (2021) 1925--1936.

\bibitem{heart}
J.~V. Firmino, M.~Melo, V.~Salemi, K.~Bringel, D.~Leone, R.~Pereira, M.~Rodrigues, Heart failure recognition using human voice analysis and artificial intelligence, Evolutionary Intelligence 16~(6) (2023) 2015--2027.

\bibitem{rev9}
J.~D.~S. Sara, E.~Maor, D.~Orbelo, R.~Gulati, L.~O. Lerman, A.~Lerman, Noninvasive voice biomarker is associated with incident coronary artery disease events at follow-up, in: Mayo Clinic Proceedings, Vol.~97, Elsevier, 2022, pp. 835--846.

\bibitem{depression}
R.~Brueckner, N.~Kwon, V.~Subramanian, N.~Blaylock, H.~O’Connell, Audio-based detection of anxiety and depression via vocal biomarkers, in: Future of Information and Communication Conference, Springer, 2024, pp. 124--141.

\bibitem{voice_3}
J.~Hlavni{\v{c}}ka, T.~Tykalov{\'a}, O.~Ulmanov{\'a}, P.~Du{\v{s}}ek, D.~Hor{\'a}kov{\'a}, E.~R{\r{u}}{\v{z}}i{\v{c}}ka, J.~Klemp{\'{i}}{\v{r}}, J.~Rusz, Characterizing vocal tremor in progressive neurological diseases via automated acoustic analyses, Clinical Neurophysiology 131~(5) (2020) 1155--1165.

\bibitem{ahmadlivoice}
N.~Ahmadli, M.~A. Sarsil, B.~Mizrak, K.~Karauzum, A.~Shaker, E.~Tulumen, D.~Mirzamidinov, D.~Ural, O.~Ergen, Voice-driven mortality prediction in hospitalized heart failure patients: A machine learning approach enhanced with diagnostic biomarkers, arXiv preprint arXiv:2402.13812 (2024).

\bibitem{mechanics}
Z.~Zhang, Mechanics of human voice production and control, The journal of the acoustical society of america 140~(4) (2016) 2614--2635.

\bibitem{glucovoicerev}
J.~Sidorova, P.~Carbonell, M.~{\v{C}}uki{\'c}, Blood glucose estimation from voice: first review of successes and challenges, Journal of Voice 36~(5) (2022) 737--e1.

\bibitem{neuropathy}
B.~C. Callaghan, G.~Gallagher, V.~Fridman, E.~L. Feldman, Diabetic neuropathy: what does the future hold?, Diabetologia 63 (2020) 891--897.

\bibitem{myopathy}
T.~P. Saliu, T.~Kumrungsee, K.~Miyata, H.~Tominaga, N.~Yazawa, K.~Hashimoto, M.~Kamesawa, N.~Yanaka, Comparative study on molecular mechanism of diabetic myopathy in two different types of streptozotocin-induced diabetic models, Life Sciences 288 (2022) 120183.

\bibitem{dys}
S.~F. Weinreb, K.~Piersiala, A.~T. Hillel, L.~M. Akst, S.~R. Best, Dysphonia and dysphagia as early manifestations of autoimmune inflammatory myopathy, American journal of otolaryngology 42~(1) (2021) 102747.

\bibitem{pho}
A.-L. Hamdan, Z.~Kurban, S.~T. Azar, Prevalence of phonatory symptoms in patients with type 2 diabetes mellitus, Acta diabetologica 50~(5) (2013) 731--736.

\bibitem{hamdan2012vocal}
A.-l. Hamdan, J.~Jabbour, J.~Nassar, I.~Dahouk, S.~T. Azar, Vocal characteristics in patients with type 2 diabetes mellitus, European Archives of Oto-Rhino-Laryngology 269 (2012) 1489--1495.

\bibitem{instrumental}
S.~Pinyopodjanard, P.~Suppakitjanusant, P.~Lomprew, N.~Kasemkosin, L.~Chailurkit, B.~Ongphiphadhanakul, Instrumental acoustic voice characteristics in adults with type 2 diabetes, Journal of Voice 35~(1) (2021) 116--121.

\bibitem{cfrd}
P.~Suppakitjanusant, N.~Kasemkosin, A.~K. Sivapiromrat, S.~Weinstein, B.~Ongphiphadhanakul, W.~R. Hunt, V.~Sueblinvong, V.~Tangpricha, Predicting glycemic control status and high blood glucose levels through voice characteristic analysis in patients with cystic fibrosis-related diabetes (cfrd), Scientific Reports 13~(1) (2023) 8617.

\bibitem{acousticgluco}
J.~M. Kaufman, A.~Thommandram, Y.~Fossat, Acoustic analysis and prediction of type 2 diabetes mellitus using smartphone-recorded voice segments, Mayo Clinic Proceedings: Digital Health 1~(4) (2023) 534--544.

\bibitem{voicediabete}
L.~Czupryniak, E.~Sielska-Badurek, A.~Niebisz, M.~Sobol, M.~Kmiecik, K.~Jedra, E.~Szymanska-Garbacz, K.~Niemczyk, 378-p: Human voice is modulated by hypoglycemia and hyperglycemia in type 1 diabetes, Diabetes 68~(Supplement\_1) (2019).

\bibitem{patent}
P.~R. Michaelis, Detection of extreme hypoglycemia or hyperglycemia based on automatic analysis of speech patterns, uS Patent 7,925,508 (Apr.~12 2011).

\bibitem{american20212}
A.~D. Association, 2. classification and diagnosis of diabetes: standards of medical care in diabetes—2021, Diabetes care 44~(Supplement\_1) (2021) S15--S33.

\bibitem{who}
W.~H. Organization, et~al., Definition and diagnosis of diabetes mellitus and intermediate hyperglycaemia: report of a who/idf consultation (2006).

\bibitem{glucose}
L.~Norton, C.~Shannon, A.~Gastaldelli, R.~A. DeFronzo, Insulin: The master regulator of glucose metabolism, Metabolism 129 (2022) 155142.

\bibitem{oralglucose}
E.~Papachatzopoulou, C.~Chatzakis, I.~Lambrinoudaki, K.~Panoulis, K.~Dinas, N.~Vlahos, A.~Sotiriadis, M.~Eleftheriades, Abnormal fasting, post-load or combined glucose values on oral glucose tolerance test and pregnancy outcomes in women with gestational diabetes mellitus, Diabetes Research and Clinical Practice 161 (2020) 108048.

\bibitem{capev}
G.~B. Kempster, B.~R. Gerratt, K.~V. Abbott, J.~Barkmeier-Kraemer, R.~E. Hillman, Consensus auditory-perceptual evaluation of voice: development of a standardized clinical protocol (2009).

\bibitem{capev_turk}
E.~{\"{O}}zcebe, F.~E. Aydinli, T.~K. Ti{\u{g}}rak, {\"O}.~{\.I}ncebay, T.~Yilmaz, Reliability and validity of the turkish version of the consensus auditory-perceptual evaluation of voice (cape-v), Journal of voice 33~(3) (2019) 382--e1.

\bibitem{disvoice1}
E.~A. Belalc{\'{a}}zar-Bolanos, J.~R. Orozco-Arroyave, J.~F. Vargas-Bonilla, T.~Haderlein, E.~N{\"o}th, Glottal flow patterns analyses for parkinson’s disease detection: acoustic and nonlinear approaches, in: International Conference on Text, Speech, and Dialogue, Springer, 2016, pp. 400--407.

\bibitem{disvoice2}
J.~C. V{\'a}squez-Correa, J.~Orozco-Arroyave, T.~Bocklet, E.~N{\"o}th, Towards an automatic evaluation of the dysarthria level of patients with parkinson's disease, Journal of communication disorders 76 (2018) 21--36.

\bibitem{disvoice3}
T.~Arias-Vergara, J.~C. V{\'a}squez-Correa, J.~R. Orozco-Arroyave, Parkinson’s disease and aging: analysis of their effect in phonation and articulation of speech, Cognitive Computation 9 (2017) 731--748.

\bibitem{disvoice4}
N.~Dehak, P.~Dumouchel, P.~Kenny, Modeling prosodic features with joint factor analysis for speaker verification, IEEE Transactions on Audio, Speech, and Language Processing 15~(7) (2007) 2095--2103.

\bibitem{curse}
N.~Altman, M.~Krzywinski, The curse (s) of dimensionality, Nat Methods 15~(6) (2018) 399--400.

\bibitem{pca1}
B.~M.~S. Hasan, A.~M. Abdulazeez, A review of principal component analysis algorithm for dimensionality reduction, Journal of Soft Computing and Data Mining 2~(1) (2021) 20--30.

\bibitem{logistic}
S.~Nusinovici, Y.~C. Tham, M.~Y.~C. Yan, D.~S.~W. Ting, J.~Li, C.~Sabanayagam, T.~Y. Wong, C.-Y. Cheng, Logistic regression was as good as machine learning for predicting major chronic diseases, Journal of clinical epidemiology 122 (2020) 56--69.

\bibitem{sigmoid}
J.~Dombi, T.~J{\'o}n{\'a}s, Generalizing the sigmoid function using continuous-valued logic, Fuzzy Sets and Systems 449 (2022) 79--99.

\bibitem{mle}
J.-X. Pan, K.-T. Fang, J.-X. Pan, K.-T. Fang, Maximum likelihood estimation, Growth curve models and statistical diagnostics (2002) 77--158.

\bibitem{loss}
Q.~Wang, Y.~Ma, K.~Zhao, Y.~Tian, A comprehensive survey of loss functions in machine learning, Annals of Data Science (2020) 1--26.

\bibitem{regularization}
Y.~Tian, Y.~Zhang, A comprehensive survey on regularization strategies in machine learning, Information Fusion 80 (2022) 146--166.

\bibitem{hyperparameter}
L.~Yang, A.~Shami, On hyperparameter optimization of machine learning algorithms: Theory and practice, Neurocomputing 415 (2020) 295--316.

\bibitem{gridsearch}
G.~Ranjan, A.~K. Verma, S.~Radhika, K-nearest neighbors and grid search cv based real time fault monitoring system for industries, in: 2019 IEEE 5th international conference for convergence in technology (I2CT), IEEE, 2019, pp. 1--5.

\bibitem{loocv}
T.-T. Wong, Performance evaluation of classification algorithms by k-fold and leave-one-out cross validation, Pattern recognition 48~(9) (2015) 2839--2846.

\end{thebibliography}



\end{document}